\documentclass[11pt,a4paper]{article}

\usepackage[utf8]{inputenc}
\usepackage[T1]{fontenc}
\usepackage[english]{babel}
\usepackage{amsmath}
\usepackage{amsfonts}
\usepackage{amssymb}
\usepackage{graphicx}
\usepackage{booktabs}
\usepackage{hyperref}
\usepackage{geometry}
\usepackage{fvextra}
\usepackage{indentfirst}
\usepackage{natbib}
\usepackage{dirtree}

\usepackage{abstract}

\title{Assessing Cognitive Effort in L2 Idiomatic Processing: An Eye-Tracking Dataset}

\author{
Eduardo Santos \and
Juliana Carvalho \and
César Rennó-Costa \\[0.8em]
\small\itshape Federal University of Rio Grande do Norte (UFRN), Brazil \\
\small\itshape contact: cesar@imd.ufrn.br
}
\date{\today}

\begin{document}

\maketitle

\begin{abstract}
This paper presents the development and validation of an eye-tracking dataset designed to investigate how second-language (L2) learners process idiomatic expressions. While native speakers often rely on direct retrieval of figurative meanings, L2 speakers frequently adopt a "literal-first" approach, which incurs measurable cognitive costs. This resource captures these costs through ocular metrics recorded from Portuguese L1 speakers of English across all CEFR proficiency levels (A1–C2). Although the study uses entry-level 60 Hz hardware (Tobii Pro Spark), we demonstrate that this sampling rate provides sufficient data density to detect macro-cognitive events such as fixations and regressions in reading. Preliminary analysis validates the dataset by revealing a strong inverse correlation between language proficiency and regressive eye movements. Integrated into the MIA (Modeling Idiomaticity in Human and Artificial Language Processing) initiative, this dataset serves as a cognitively grounded benchmark for evaluating both human processing models and the alignment of large language models with human-like figurative understanding.

\end{abstract}


\section{Introduction}

The processing of Multi-Word Expressions (MWEs), and specifically idiomatic expressions, remains a persistent challenge in both cognitive science and computational linguistics. Idioms such as "spill the beans" or "break the ice" represent a semantic paradox: they are syntactically comprised of individual lexical items that retain their standard grammatical properties, yet their semantic payload is non-compositional, meaning the aggregate sense cannot be derived solely from the summation of the constituent parts \citep{sag_multiword_2002}. For a native speaker, accessing these meanings is often instantaneous, facilitated by what psycholinguistic models describe as a direct access route, in which the idiom is retrieved as a single lexical unit \citep{swinney_access_1979}. However, for a second-language (L2) learner, these expressions constitute a significant central challenge to comprehension, often triggering a breakdown in automatic processing and necessitating resource-intensive cognitive reanalysis \citep{sag_multiword_2002,siyanova2011}.


The urgency of creating robust resources to study this phenomenon is underscored by the limitations of current Natural Language Processing (NLP) architectures. Although large language models (LLMs) have achieved remarkable fluency, they continue to struggle with the nuances of figurative language and often operate as 'black boxes' regarding their decision-making processes. The 'Modeling Idiomaticity in Human and Artificial Language Processing' (MIA) initiative was established to address these precise limitations by integrating insights from human cognitive processing into the design and evaluation of artificial systems \citep{Villavicencio2020MIA}. While our work uses eye-tracking to capture the real-time cognitive cost incurred by human learners, complementary approaches such as the SLIME methodology employ Integrated Gradients to provide linguistic insights into how models prioritize lexical components during classification \citep{ribeiro_methodology_2024}. Both strategies aim to address the limitations of current architectures by integrating insights from human processing into the design and evaluation of robust, human-aligned AI models.

In this work, we describe the creation of a specialized dataset that captures this cognitive effort through eye-tracking. Unlike offline measures of comprehension, such as questionnaires or cloze tests, eye-tracking provides an online, millisecond-by-millisecond record of the reading process. It allows researchers to distinguish between initial lexical access (as measured by fixation durations) and late-stage syntactic or semantic integration (as measured primarily by regressive saccades). This granularity is essential for testing the "re-analysis hypothesis," which posits that L2 speakers, particularly those at lower proficiency levels, will initially parse an idiom literally before encountering a semantic violation that forces a regression and re-interpretation.

The dataset described herein was developed in collaboration with the AdMIRe (Advancing Multimodal Idiomaticity Representation) initiative and the University of Sheffield, aligning with the objectives of the SemEval-2025 Task 1.1 \citep{pickard2025}. By recording the gaze patterns of Brazilian Portuguese learners of English, systematically stratified across proficiency levels (A1–C2), this resource enables the analysis of how idiomatic processing strategies evolve throughout language acquisition. 

The remainder of this paper is structured as follows. Section~\ref{background} presents the theoretical foundations underlying idiom processing, followed by Section~\ref{related}, which situates this work within the existing literature. Section~\ref{experiment} describes the experimental design and used stimuli, while Section~\ref{methodology} details the data acquisition procedures. Section~\ref{data} outlines the data processing pipeline, and Section~\ref{dataset} describes the dataset's organization and structure. Finally, Section~\ref{analisys} presents the analyses conducted to validate the dataset and demonstrate its utility.

\section{Theoretical Background}
\label{background}

The theoretical landscape of idiom comprehension is dominated by the tension between compositional and non-compositional processing. Early models of idiom processing posited a strict dichotomy between lexical and compositional accounts. \citep{bobrow1973} provided evidence for the existence of distinct processing modes, suggesting that idiomatic expressions can be understood via a dedicated mechanism in which they are treated as unified lexical entities, akin to long words retrieved from an idiom-specific mental lexicon. In contrast, literal language was assumed to be processed compositionally, through the sequential integration of individual word meanings into a coherent semantic representation. This dichotomy laid the foundation for subsequent debates on whether idioms are accessed directly or derived through standard syntactic and semantic parsing mechanisms.

Importantly, the evolution from early to contemporary models does not fundamentally alter the nature of compositional processing itself, but rather reconceptualizes how it interacts with idiomatic meaning. While early models assumed mutually exclusive processing routes, modern frameworks posit parallel, competitive activation of both literal and figurative interpretations. This parallel processing is hypothesized to be supported by neurobiological mechanisms where the brain maintains competing interpretations through specific oscillatory patterns, a framework that integrates AI-driven modeling with neurobiological data \citep{idiart2019}. \citet{titone2014} proposed a hybrid framework in which the literal and figurative meanings of an idiom are activated in parallel, with the speed of resolution determined by factors such as familiarity and context. In native speakers, this parallel activation is heavily weighted toward the figurative meaning due to high exposure and strong lexical automaticity. This results in the well-documented "idiom superiority effect", where idioms are often processed faster than novel literal phrases of comparable length and complexity.

However, the mechanism differs fundamentally for L2 speakers. Literal Salience Resonant Hypothesis is the foundational theory guiding the design of this dataset \citep{cieslicka2006}. Cieślicka argues that for L2 learners, the literal meanings of constituent words (e.g., "spill" and "beans") are more salient and more easily accessible than the figurative meaning of the whole phrase. This is because the individual words are encountered earlier in the acquisition process and typically possess higher individual frequencies than the idiom itself. Consequently, L2 processing is characterized by a "literal-first" bias, where learners initially attempt to build meaning compositionally. Only when this compositional path fails to yield a coherent interpretation within the context do they suppress the literal meaning and seek a figurative alternative.

\section{Related Work}
\label{related}

Eye-tracking has been widely used to investigate real-time language processing, as it provides fine-grained temporal evidence of how readers allocate attention during comprehension. Metrics such as fixation duration, gaze duration, and regressions allow researchers to dissociate early lexical access from later stages of syntactic and semantic integration. In the context of idiomatic expressions, regressions are particularly informative, as they signal moments of processing difficulty and re-analysis.

Previous studies have leveraged eye-tracking to examine idiom processing in both L1 and L2 speakers. Findings suggest that native speakers often benefit from a processing advantage when encountering idiomatic expressions, whereas L2 learners tend to exhibit increased cognitive cost, frequently reflected in longer reading times and a higher number of regressions \citep{conklin2008, siyanova2011}. However, this processing advantage is not restricted to idioms alone; recent eye-tracking and self-paced reading studies indicate that highly frequent novel phrases can also trigger direct retrieval mechanisms in a manner similar to idiomatic expressions \citep{rambelli_are_2023}. These results provide empirical support for theoretical accounts such as the Literal Salience Hypothesis, which predicts a reliance on compositional processing in L2 comprehension. However, existing studies are typically limited by small sample sizes and homogeneous participant groups, restricting their ability to capture variability across proficiency levels. Moreover, few resources provide large-scale, structured datasets that link fine-grained eye-movement data to idiomatic processing in a way that is suitable for both psycholinguistic and computational analysis. Recent benchmarks, such as the Noun Compound Idiomaticity Minimal Pairs (NCIMP) dataset, have begun to fill this gap by providing 32,200 sentences in English and Portuguese, specifically designed to test model sensitivity to idiomatic shifts through minimal pair analysis \citep{he_investigating_2025}.

The present work addresses this gap by introducing a stratified eye-tracking dataset spanning the full CEFR proficiency range (A1–C2), enabling systematic investigation of how idiomatic processing strategies evolve during second-language acquisition.

Beyond psycholinguistic research, eye-tracking has gained relevance in computational linguistics as a source of cognitively grounded supervision signals. The MIA project aims to incorporate human processing data into the development of NLP systems, addressing known limitations of large language models in handling figurative language. By providing aligned gaze and linguistic data, this dataset offers a resource for bridging human cognitive behavior and machine learning models. Specifically, the cognitive friction observed in human learners provides a ground-truth signal for training AI to recognize non-compositional shifts—a process formally grounded in the Literal Salience Resonant Hypothesis. In this context, the AdMIRe shared task (SemEval-2025 Task 1) introduces multimodal benchmarks for idiomaticity. Eye-tracking data can serve as a proxy for human attention, revealing which parts of an expression are most informative and how meaning is incrementally constructed. By providing aligned gaze and linguistic data, the dataset introduced in this work offers a resource for bridging human cognitive behavior and machine learning models.

\section{Experiment}
\label{experiment}
\subsection{Design and Stimuli}
The experimental protocol was designed to rigorously test the hypothesis that ocular behavior during idiom processing differs significantly across proficiency levels in L2 speakers. The study replicated and adapted an international protocol to the local context of Brazilian Portuguese-speaking learners of English.

The core of the dataset is the "Potentially Idiomatic Expressions" (PIE) Context Data. The term "Potentially Idiomatic" is crucial: it refers to strings of words that can function as idioms but can also function literally depending on the context (e.g., "skating on thin ice"). The items were drawn primarily from the MAGPIE corpus (\citep{haagsma2020}), a large, sense-annotated corpus derived from the British National Corpus (BNC). MAGPIE is unique in that it includes annotations for both idiomatic and literal occurrences of the same word strings, making it the ideal source for creating a contrastive dataset. Additional items were sourced from SemEval shared task datasets to ensure alignment with community benchmarks.

The dataset employs a matched-pairs design. For each target MWE (e.g., "spill the beans"), participants were presented with full sentences. The critical manipulation was the context, which biased the interpretation toward either the figurative or the literal sense.

\begin{itemize}
    \item \textbf{Figurative Context:} ``After hours of interrogation, the suspect finally decided to spill the beans about the robbery.''
    
    \item \textbf{Literal Context:} ``The clumsy chef managed to spill the beans all over the kitchen floor.''
\end{itemize}

This design allows researchers to subtract the reading time associated with the literal words (which is constant across conditions) and isolate the cognitive cost of the figurative meaning. The text was presented in Open Sans, centered on the screen, with a font size relative to the screen height to ensure legibility and a standardized character width for gaze mapping.

An example of the stimulus presentation interface is shown in Figure~\ref{fig:stimulus}. The figure illustrates how sentences were displayed to participants during the experiment, including text positioning and overall layout.

\begin{figure}[h]\label{fig:stimulus}
    \centering
    \includegraphics[width=1\linewidth]{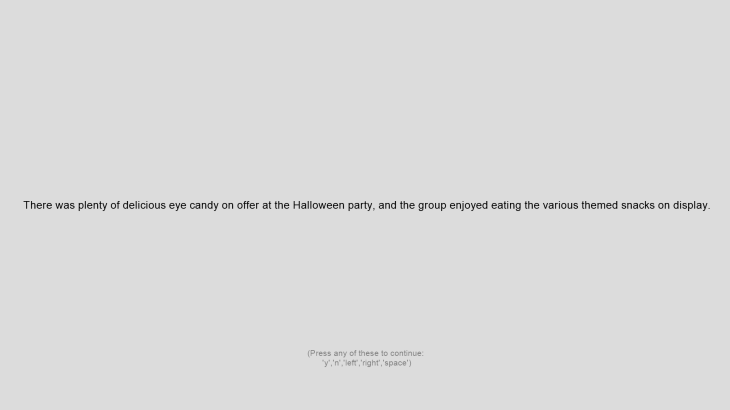}
    \caption{Example of stimulus presentation during the experiment, showing sentence layout and visual configuration.}
    \label{fig:stimulus}
\end{figure}
A defining feature of this resource is the stratification of participants. The target population consisted of university students, predominantly from the Federal University of Rio Grande do Norte (UFRN) and the Digital Metropolis Institute (IMD). All participants were native speakers of Brazilian Portuguese (L1) and non-native speakers of English (L2).
\linebreak
\subsection{Proficiency Assessment (CEFR)}

Participants were categorized based on the Common European Framework of Reference for Languages (CEFR), a widely accepted standard for grading language proficiency. The levels are defined as follows:

\begin{itemize}
    \item \textbf{A1/A2 (Basic User):} Can understand sentences and frequently used expressions related to areas of immediate relevance.
    
    \item \textbf{B1/B2 (Independent User):} Can understand the main points of clear standard input. B2 users can interact with sufficient fluency to make regular interaction with native speakers quite possible.
    
    \item \textbf{C1/C2 (Proficient User):} Can understand with ease virtually everything heard or read. C2 users can express themselves spontaneously, fluently, and precisely, differentiating finer shades of meaning even in complex situations.
\end{itemize}

Proficiency was determined via a structured self-report form, validated against university placement standards. This metadata was encoded into a CSV file (\texttt{group\_metrics}) linked to the eye-tracking logs, enabling the analysis of gaze behavior as a continuous function of proficiency. The hypothesis posits a transition from ``decoding'' behaviors (a high number of regressions) in A1/B1 speakers to ``direct access'' behaviors (a low number of regressions) in C1/C2 speakers.

The data collection was conducted in accordance with ethical standards for human subject research. The protocol was reviewed by the relevant institutional review board (Comitê de Ética em Pesquisa at UFRN) to ensure informed consent and data anonymization. The "Comitê de Ética em Pesquisa" (CEP) operates under the regulations of the Brazilian National Health Council (CONEP) and requires strict adherence to privacy and participant safety protocols. All participant IDs in the dataset are anonymized to protect privacy.

\section{Methodology and Data Acquisition}
\label{methodology}

Data collection took place in a controlled laboratory environment at the Digital Metropolis Institute (IMD/UFRN). As the setting was an adapted acquisition station rather than a shielded eye-tracking facility, strict physical protocols were implemented to ensure data integrity. Ambient lighting was controlled to minimize glare and enable consistent pupil detection, while the eye tracker was rigidly mounted to the display to prevent hardware shifts. To maintain a constant visual angle, the viewing distance was standardized, and head stabilization was required to compensate for the tracking hardware's single-camera architecture.

\begin{figure}[h]
    \centering
    \includegraphics[width=0.5\linewidth]{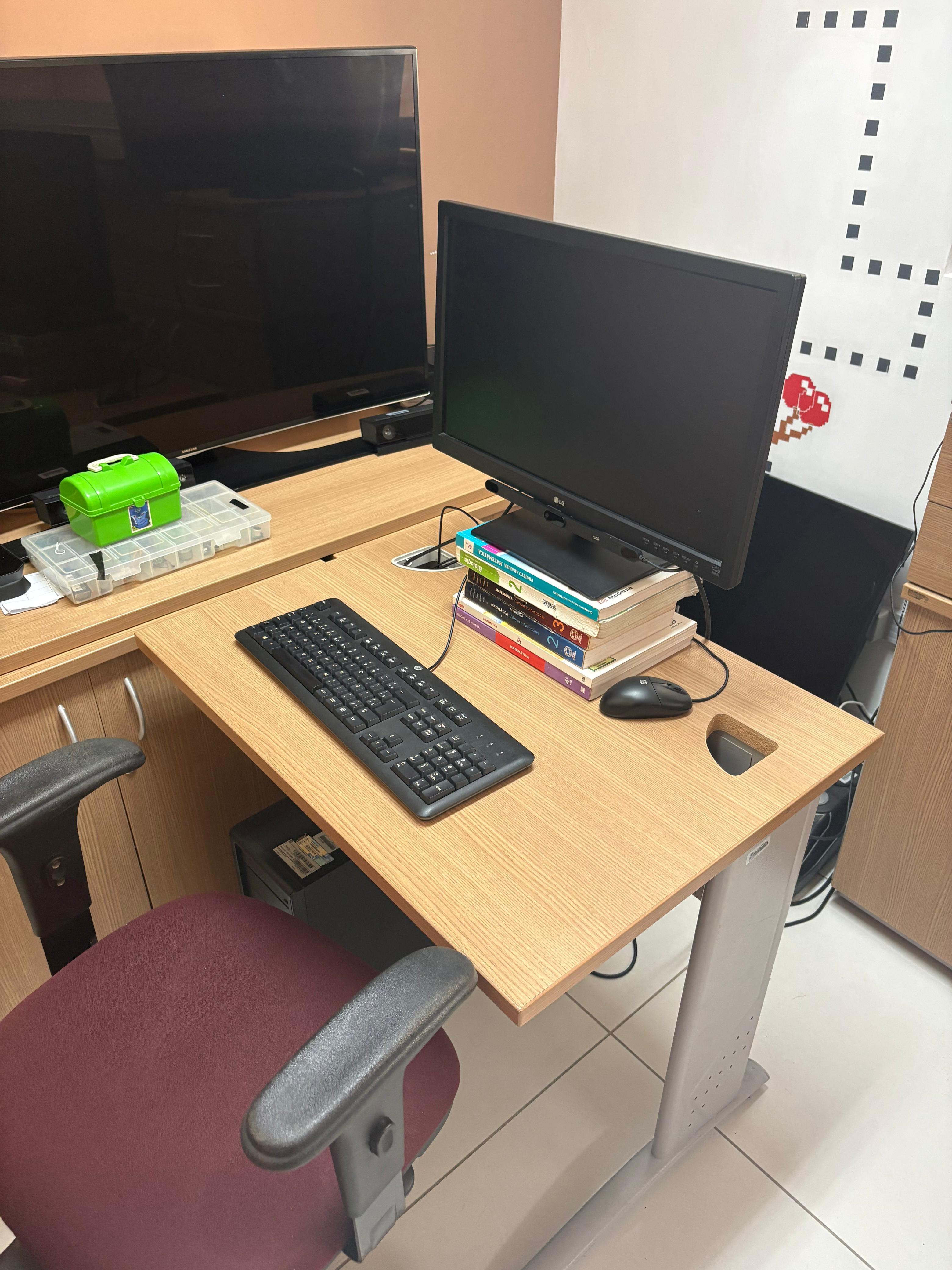}
    \caption{Experimental setup showing participant seating position, display, and eye-tracking device.}
    \label{fig:setup}
\end{figure}

Figure~\ref{fig:setup} illustrates the experimental setup. Participants were seated in a fixed chair positioned directly in front of a desktop display, with the eye tracker mounted below the screen. The configuration was designed to minimize head movement and ensure consistent alignment between the participant's gaze and the stimulus presentation area.

The study used a Tobii Pro Spark, a screen-based eye tracker operating at a 60 Hz sampling rate, yielding a temporal resolution of approximately 16.67 ms. Although this frequency is often categorized as entry-level hardware, it provides the necessary data density for the macro-cognitive scope of this study. Given that human reading fixations typically last at least 100 ms, a 60 Hz sampling rate yields multiple data points per fixation, ensuring sufficient robustness for fixation detection algorithms.

Although the device records binocular gaze data, it lacks the onboard event-detection algorithms typically found in higher-frequency models. Consequently, raw gaze coordinates were exported for offline processing, requiring a custom pipeline to identify fixations and saccades directly from the raw telemetry.

The experimental environment was governed by PsychoPy \citep{peirce2007}, which managed stimulus presentation and data logging. The protocol followed a fixed sequence, beginning with a mandatory 9-point calibration to map gaze vectors to screen coordinates. Participants then completed a practice phase comprising six sentences; these data were stored in a separate pre-experiment directory and excluded from the final analysis.
During the main task, target sentences containing Potentially Idiomatic Expressions (PIEs) were presented, interspersed with comprehension checks every four sentences. These numeric or scalar inquiries were randomized and logged automatically to ensure sustained participant engagement throughout the session. To mitigate fatigue effects, participants were given a scheduled break midway through the experiment.

\section{Data Processing }
\label{data}
A key contribution of this project is a transparent, open-source pipeline that transforms raw hardware output into psycholinguistic metrics. This pipeline addresses the "missing link" between the proprietary data formats of hardware vendors and the input requirements of analysis libraries like PyGaze (\citep{dalmaijer2014}), which was used in this work.

\subsection{From Raw Data to Processed Data }
The raw ocular output is archived in .csv and .psydata formats within a hierarchical directory structure to maintain data provenance. These files contain high-resolution, longitudinal time-series data consisting of raw spatial coordinates. The primary data streams include Gaze Point left eye and Gaze Point right eye, along with the corresponding sentence. At a sampling frequency of 60 Hz, the system yields a temporal resolution of approximately $16.67\,\text{ms}$ per data point, resulting in $120$ observations per second across both eyes. While this granular telemetry provides the requisite density for algorithmic event detection, these raw oculomotor signals are pre-cognitive and must undergo significant transformation, specifically the identification of fixations and saccades, before they can be mapped to high-level psycholinguistic processing.

To facilitate integration with the PyGaze library, the raw oculomotor data were preprocessed through a series of deterministic transformations. Initially, the gaze data, stored as tuple-valued entries for the left and right eyes, were decomposed into separate columns representing the horizontal (x) and vertical (y) coordinates. Subsequently, these coordinates were converted from normalized space to pixel units relative to the stimulus display. Finally, a cyclopean gaze point was computed by averaging the corresponding x- and y-coordinates from both eyes, yielding a single unified gaze signal for downstream analysis. Ocular artifacts, such as blinks or transient tracking loss, were addressed by interpolating minor data gaps, whereas significant signal dropout led to trial exclusion. 

\subsection{Event Detection }
Cognitive events were derived using the Identification by Dispersion Threshold (I-DT) algorithm, which identifies fixations by applying a moving window, parameterized by a minimum duration, to detect clusters in which spatial dispersion remains below a $1^\circ$ visual angle threshold. These parameters were calibrated in the PyGaze environment to match the 60 Hz sampling frequency and the display geometry of the IMD laboratory.

To rigorously evaluate the re-analysis hypothesis, the experimental pipeline focused on quantifying regressions, defined as saccadic movements directed toward previously encoded lexical regions (Salvucci and Goldberg, 2000). Given that PyGaze’s native saccade-detection algorithms typically rely on velocity thresholds and are less robust when applied to 60Hz telemetry, a custom heuristic was developed to infer regressions from fixation sequences. This algorithm classifies an event as a regression if the spatial coordinates of $Fixation_{n}$ exhibit an X-coordinate significantly lower than that of $Fixation_{n-1}$ ($X_{n} < X_{n-1}$), provided the Y-coordinate remains within the established vertical boundaries of the current line of text. By prioritizing the sequential logic of fixation positions over raw saccadic velocity, this methodology circumvented the hardware's signal-to-noise limitations and provided a direct, reliable proxy for the cognitive labor involved in semantic re-evaluation.

To ensure the reliability of the proposed regression-detection heuristic, the extracted fixation sequences were manually inspected. Table~\ref{tab:regression_sanity} presents a representative subset of detected events, illustrating cases in which the horizontal displacement between consecutive fixations satisfies the condition $X_n < X_{n-1}$ while remaining within the same textual line. This qualitative sanity check confirms that the algorithm correctly identifies backward eye movements consistent with established definitions of reading regressions.

\begin{table}[h]
\centering
\caption{Sample of detected regression events. Time is expressed in samples (60 Hz), where each unit corresponds to one sample, and regressions correspond to negative horizontal displacement ($dx < 0$).}
\label{tab:regression_sanity}

\resizebox{\linewidth}{!}{
\begin{tabular}{cccccccc}
\toprule
start\_time & end\_time & duration & x & y & prev\_x & dx & is\_regression \\
\midrule
24  & 49  & 25 & 131.109 & 502.839 & 216.674 & -85.565  & True \\
67  & 148 & 81 & 118.926 & 513.145 & 131.109 & -12.183  & True \\
201 & 221 & 20 & 298.927 & 520.598 & 370.683 & -71.756  & True \\
273 & 284 & 11 & 321.841 & 538.352 & 475.635 & -153.794 & True \\
286 & 294 & 8  & 225.181 & 532.255 & 321.841 & -96.660  & True \\
\bottomrule
\end{tabular}
}
\end{table}

\section{Dataset Organization and Content}
\label{dataset}

\begin{figure}[h]
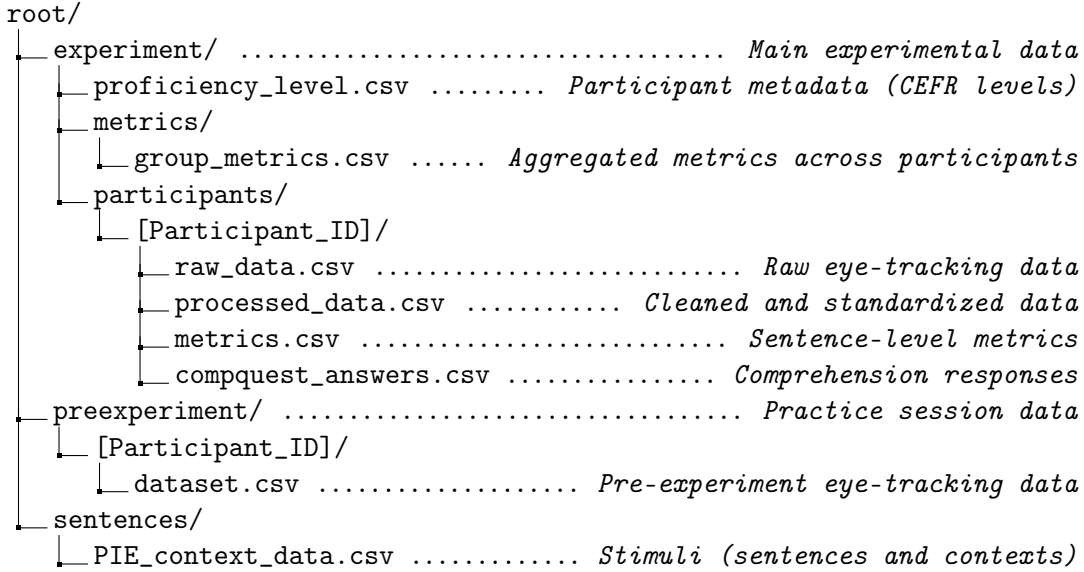

\dirtree{%
.1 root/.
.2 experiment/ \dotfill{} \textit{Main experimental data}.
.3 proficiency\_level.csv \dotfill{} \textit{Participant metadata (CEFR levels)}.
.3 metrics/.
.4 group\_metrics.csv \dotfill{} \textit{Aggregated metrics across participants}.
.3 participants/.
.4 [Participant\_ID]/.
.5 raw\_data.csv \dotfill{} \textit{Raw eye-tracking data}.
.5 processed\_data.csv \dotfill{} \textit{Cleaned and standardized data}.
.5 metrics.csv \dotfill{} \textit{Sentence-level metrics}.
.5 compquest\_answers.csv \dotfill{} \textit{Comprehension responses}.
.2 preexperiment/ \dotfill{} \textit{Practice session data}.
.3 [Participant\_ID]/.
.4 dataset.csv \dotfill{} \textit{Pre-experiment eye-tracking data}.
.2 sentences/.
.3 PIE\_context\_data.csv \dotfill{} \textit{Stimuli (sentences and contexts)}.
}
\caption{Overview of the eye-tracking dataset directory structure.}
    \label{fig:data_structure}
\end{figure}

The finalized dataset is organized within a hierarchical file structure designed to support computational reproducibility and transparency (Figure \ref{fig:data_structure}). This organization enforces a clear data lineage, allowing researchers to perform bidirectional tracing across different levels of analysis, from high-level aggregated metrics to participant-specific observations and, ultimately, to the original raw gaze coordinates. By explicitly separating raw data, processed outputs, and derived metrics, the dataset facilitates independent verification of each stage of the processing pipeline, enabling researchers to reproduce results or apply alternative analytical methods without loss of provenance. This structure also supports multi-level analysis, enabling investigation of phenomena at the group level (e.g., proficiency-based comparisons), the participant level, and the sentence level.

At the root level, the dataset is divided into three main directories: \textit{experiment}, \textit{preexperiment}, and \textit{sentences}. The \textit{sentences} directory contains the file \textit{PIE\_context\_data.csv}, which stores the textual stimuli used in the experiment, including both literal and figurative contexts for each target expression. The \textit{experiment/} directory contains the core experimental data. At this level, the file \textit{proficiency\_level.csv} provides participant metadata, including anonymized identifiers and their corresponding CEFR proficiency levels. Aggregated metrics across all participants are stored in \textit{metrics/group\_metrics.csv}, which includes summary measures such as total fixations and total regressions.

Individual-level data are organized within the \textit{participants/} subdirectory, where each participant is assigned a dedicated folder identified by a unique ID. Within each participant folder, the file \textit{raw\_data.csv} contains the original temporal-spatial gaze data recorded by the eye tracker, while \textit{processed\_data.csv} provides a cleaned and standardized version suitable for analysis. The file \textit{metrics.csv} stores derived measures at the sentence level, such as fixation counts and regression counts, and \textit{compquest\_answers.csv} logs responses to comprehension questions.

The \textit{preexperiment/} directory mirrors this structure for the practice phase of the experiment. It contains participant-specific folders with \textit{dataset.csv} files corresponding to pre-experimental trials, which can be used for calibration, baseline analysis, or quality control, but are not included in the main experimental results.

\section{Analisys and Validation}
\label{analisys}
To validate the dataset and processing pipeline, a preliminary analysis examined the relationship between language proficiency (CEFR) and ocular behaviors (Regressions and Fixations). The results provide strong support for the validity of the data and the underlying psycholinguistic hypotheses.

The analysis of regression counts reveals a stark contrast between novice and expert L2 readers. As proficiency increases, the number of regressions decreases, and the behavior becomes more consistent (with a lower standard deviation).

\begin{table}[h!]
\centering
\caption{Descriptive Statistics of Ocular Regressions by Proficiency Level}
\label{tab:results}
\begin{tabular}{lcccccc}
\toprule
\textbf{Proficiency} & \textbf{Mean} & \textbf{Median} & \textbf{SD} & \textbf{Min} & \textbf{Max} \\ \midrule
B1 (Intermediate) & 1,776 & 1,240 & $\pm 1,297$ & 976 & 4,046 \\
B2 (Upper Int.) & 1,439 & 1,271 & $\pm 546$ & 787 & 2,052 \\
C1 (Advanced) & 1,103 & 1,240 & $\pm 690$ & 355 & 1,715 \\
C2 (Proficient) & 987 & 962 & $\pm 115$ & 887 & 1,114 \\ \bottomrule
\end{tabular}
\end{table}

The B1 cohort exhibits marked heterogeneity, with peak counts of $4{,}046$ regressions per participant, indicating acute cognitive demand during lexical integration. Within the framework of the \textit{Literal Salience Hypothesis}, this behavior indicates a recursive processing loop: the participant likely executes a literal bottom-up decoding, encounters a semantic mismatch at the sentence terminus, and initiates compensatory regressions to the idiomatic constituent for re-evaluation. The substantial standard deviation within this group suggests a wide proficiency range, from emerging idiomatic competence to reliance on literal decomposition, which aligns with the characterization of B1 readers as exhibiting high-effort, often chaotic reading behavior. At the B2 level, regression counts decrease, indicating a transition phase in which readers begin to stabilize their processing strategies. However, the still-moderate variability suggests that consistency has not yet been fully achieved, reflecting a developmental zone between effortful decoding and more efficient comprehension. For C1 learners, the reduction in regressions becomes more pronounced, signaling the emergence of fluent processing. This stage is characterized by increased lexical and idiomatic familiarity, allowing readers to rely less on re-analysis and more on predictive and integrative mechanisms during reading. Finally, the C2 group exhibits low regression counts and minimal variance, indicating direct lexical access and native-like stability. The convergence of behavior within this group indicates that proficient users have adopted highly efficient reading strategies, minimizing the need for regressions. Furthermore, a robust positive correlation exists between \textit{Total Fixations} and \textit{Total Regressions}; B1 learners predominantly occupy the upper-right quadrant of the distribution (high-frequency fixations and regressions), a signature of labored processing, whereas C2 learners cluster in the lower-left quadrant, reflecting automated, efficient processing. This systematic correlation validates regressions as a reliable index of cognitive workload rather than stochastic noise, establishing the C2 group as a stable baseline for target L2 processing in computational linguistic models.

\section{Data Availability and Usage}

In alignment with the principles of Open Science and the Language Resources and Evaluation mandate for reproducibility, this dataset and the associated code tools are made available to the research community. The full dataset, including raw logs, processed files, and the \texttt{group\_metrics} summary, is hosted on the ORDA (Online Research Data) repository of the University of Sheffield, associated with the AdMIRe SemEval-2025 task.

\noindent\textbf{Dataset DOI:} \url{https://doi.org/10.5281/zenodo.19582953}

\noindent\textbf{License:} Creative Commons Attribution 4.0 International (CC-BY 4.0), allowing reuse and adaptation with appropriate credit.
The custom Python scripts used for the PsychoPy experiment and the post-processing pipeline (including the regression detection algorithm) are available to facilitate replication or adaptation for other language pairs.
GitHub Repository: Code and sample data are linked via the SemEval-2025 Task 1 website.

\section{Conclusion}

Understanding how second-language (L2) learners process idiomatic expressions remains a central challenge at the intersection of psycholinguistics and computational linguistics. Idiomaticity exposes fundamental limitations in both human language acquisition and artificial language models, particularly due to its non-compositional and context-dependent nature. Addressing this challenge requires resources that capture not only the outcome of comprehension, but the underlying cognitive processes that unfold in real time.

In this work, we introduce an eye-tracking dataset designed to capture the cognitive effort required to process L2 idiomatic expressions. Using high-resolution gaze data, the dataset provides detailed information on fixation patterns and regressive movements, enabling investigation of the re-analysis hypothesis. The resource is further strengthened by its stratification by CEFR proficiency levels, enabling analysis of how processing strategies evolve throughout language acquisition. The main contributions of this work are threefold: (i) the creation of a structured and scalable eye-tracking dataset capturing fine-grained cognitive signals during idiom comprehension; (ii) the development of a transparent and reproducible data processing pipeline bridging raw eye-tracking data and psycholinguistic metrics; and (iii) an initial validation demonstrating a clear relationship between language proficiency and regression behavior. In addition, the dataset is organized to ensure full traceability across different levels of analysis, supporting both psycholinguistic investigation and computational modeling.

Despite these contributions, some limitations must be acknowledged. The dataset is constrained by the use of entry-level eye-tracking hardware operating at 60 Hz, which limits the temporal precision of event detection compared to higher-frequency systems. Also, although the participant pool is stratified by proficiency levels, it is restricted to a specific L1 population (Brazilian Portuguese speakers), which may limit generalizability to other linguistic backgrounds. Future work will focus primarily on expanding the dataset by increasing the number of participants and observations, improving statistical robustness, and enabling more fine-grained analyses. In addition, further work will explore more detailed analyses of the existing data, including deeper investigation of fixation patterns, regression dynamics, and their relationship with proficiency levels. These efforts aim to strengthen the empirical foundation of the dataset and support more comprehensive studies of cognitive effort during idiomatic processing.

To support reproducibility and facilitate adoption, detailed documentation of the data collection protocol is available online.\footnote{Data Collection Tutorial: \url{https://hackmd.io/lLzZ4YC7QPm2TnW3ax9Pgg}}

\section*{Acknowledgements}
This work was supported by Royal Society and the Newton Fund (NAF/R2/202209), the INCT-NeuroComp (CNPq 408389/2024-9), the Advanced Knowledge Center in Immersive Technologies (AKCIT), with financial resources from the PPI IoT of the MCTI grant number 057/2023, signed with EMBRAPII, the Fundação de Amparo à Pesquisa do Estado de Goiás (FAPEG 64448878/2024), and by CAPES-PrInt (Coordenação de Aperfeiçoamento de Pessoal de Nível Superior) in Brazil. We acknowledge the support of the Digital Metropolis Institute (IMD/UFRN) and NPAD/UFRN for providing the laboratory infrastructure and the students who participated in the study. The AdMIRe shared task organizers and the University of Sheffield NLP group - Thomas Pickard, Marco Idiart and Aline Villavicencio - contributed significantly to the experimental design.

\bibliographystyle{plainnat}
\bibliography{references}

@article{bobrow1973,
  title={On catching on to idiomatic expressions},
  author={Bobrow, Samuel A. and Bell, Susan M.},
  journal={Memory \& Cognition},
  volume={1},
  number={3},
  pages={343--346},
  year={1973}
}

@article{titone2014,
  title={Toward a compositional view of idiom processing},
  author={Titone, Debra A. and Libben, Gary},
  journal={Journal of Experimental Psychology: General},
  year={2014}
}

@article{cieslicka2006,
  title={Literal salience in on-line processing of idiomatic expressions by second language learners},
  author={Cie{\'s}licka, Anna},
  journal={Second Language Research},
  volume={22},
  number={2},
  pages={115--144},
  year={2006}
}

@article{conklin2008,
  title={Formulaic sequences: Are they processed more quickly than nonformulaic language by native and nonnative speakers?},
  author={Conklin, Kathy and Schmitt, Norbert},
  journal={Applied Linguistics},
  volume={29},
  number={1},
  pages={72--89},
  year={2008}
}

@article{siyanova2011,
  title={Adding more fuel to the fire: An eye-tracking study of idiom processing by native and non-native speakers},
  author={Siyanova-Chanturia, Anna and Conklin, Kathy and Schmitt, Norbert},
  journal={Studies in Second Language Acquisition},
  volume={33},
  number={2},
  pages={323--346},
  year={2011}
}

@article{haagsma2020,
  title={MAGPIE: A large corpus of potentially idiomatic expressions},
  author={Haagsma, Hessel and Nissim, Malvina and Bos, Johan},
  journal={Proceedings of LREC},
  year={2020}
}

@article{pickard2025,
  title={SemEval-2025 Task 1: Advancing Multimodal Idiomaticity Representation (AdMIRe)},
  author={Pickard, Thomas et al.},
  journal={Proceedings of SemEval},
  year={2025}
}

@article{peirce2007,
  title={PsychoPy—Psychophysics software in Python},
  author={Peirce, Jonathan W.},
  journal={Journal of Neuroscience Methods},
  volume={162},
  number={1-2},
  pages={8--13},
  year={2007}
}

@article{dalmaijer2014,
  title={PyGaze: An open-source, cross-platform toolbox for minimal-effort programming of eyetracking experiments},
  author={Dalmaijer, Edwin S. and Math{\^o}t, Sebastiaan and Van der Stigchel, Stefan},
  journal={Behavior Research Methods},
  volume={46},
  number={4},
  pages={913--921},
  year={2014}
}

@inproceedings{rambelli_are_2023,
	location = {Dubrovnik, Croatia},
	title = {Are Frequent Phrases Directly Retrieved like Idioms? An Investigation with Self-paced Reading and Language Models},
	url = {https://hal.science/hal-04098473},
	shorttitle = {Are Frequent Phrases Directly Retrieved like Idioms?},
	booktitle = {Workshop on Multiword Expressions ({MWE} 2023)},
	author = {Rambelli, Giulia and Chersoni, Emmanuele and Senaldi, Marco and Blache, Philippe and Lenci, Alessandro},
	urldate = {2026-04-14},
	date = {2023-05},
    year = {2023},
}

@article{idiart2019,
	title = {How the Brain Represents Language and Answers Questions? Using an {AI} System to Understand the Underlying Neurobiological Mechanisms},
	volume = {13},
	issn = {1662-5188},
	url = {https://www.frontiersin.org/articles/10.3389/fncom.2019.00012},
	shorttitle = {How the Brain Represents Language and Answers Questions?},
	journal = {Frontiers in Computational Neuroscience},
	author = {Idiart, Marco A. P. and Villavicencio, Aline and Katz, Boris and Rennó-Costa, César and Lisman, John},
	urldate = {2023-05-31},
	year = {2019},
}

@article{he_investigating_2025,
	title = {Investigating Idiomaticity in Word Representations},
	volume = {51},
	issn = {0891-2017},
	url = {https://doi.org/10.1162/coli_a_00546},
	doi = {10.1162/coli_a_00546},
	pages = {505--555},
	number = {2},
	journal = {Computational Linguistics},
	author = {He, Wei and Vieira, Tiago Kramer and Garcia, Marcos and Scarton, Carolina and Idiart, Marco and Villavicencio, Aline},
	urldate = {2026-04-14},
	date = {2025-06-24},
    year = {2025},
}

@misc{ribeiro_methodology_2024,
	title = {A Methodology for Explainable Large Language Models with Integrated Gradients and Linguistic Analysis in Text Classification},
	url = {http://arxiv.org/abs/2410.00250},
	doi = {10.48550/arXiv.2410.00250},
	number = {{arXiv}:2410.00250},
	publisher = {{arXiv}},
	author = {Ribeiro, Marina and Malcorra, Bárbara and Mota, Natália B. and Wilkens, Rodrigo and Villavicencio, Aline and Hubner, Lilian C. and Rennó-Costa, César},
	urldate = {2025-01-06},
	date = {2024-09-30},
    year = {2024},
	eprinttype = {arxiv},
	eprint = {2410.00250 [cs]},
}

@article{swinney_access_1979,
	title = {The access and processing of idiomatic expressions},
	volume = {18},
	issn = {0022-5371},
	url = {https://www.sciencedirect.com/science/article/pii/S0022537179902846},
	doi = {10.1016/S0022-5371(79)90284-6},
	pages = {523--534},
	number = {5},
	journal = {Journal of Verbal Learning and Verbal Behavior},
	shortjournal = {Journal of Verbal Learning and Verbal Behavior},
	author = {Swinney, David A. and Cutler, Anne},
	urldate = {2026-04-14},
	date = {1979-10-01},
    year = {1979},
}

@inproceedings{sag_multiword_2002,
	location = {Berlin, Heidelberg},
	title = {Multiword Expressions: A Pain in the Neck for {NLP}},
	isbn = {978-3-540-45715-2},
	doi = {10.1007/3-540-45715-1_1},
	shorttitle = {Multiword Expressions},
	pages = {1--15},
	booktitle = {Computational Linguistics and Intelligent Text Processing},
	publisher = {Springer},
	author = {Sag, Ivan A. and Baldwin, Timothy and Bond, Francis and Copestake, Ann and Flickinger, Dan},
	editor = {Gelbukh, Alexander},
	year = {2002},
	langid = {english},
}

@misc{Villavicencio2020MIA,
  author       = {Villavicencio, Aline},
  title        = {{MIA: Modeling Idiomaticity in Human and Artificial Language Processing}},
  howpublished = {UK Research and Innovation (UKRI) Grant EP/T02450X/1},
  year         = {2020},
  url          = {https://gtr.ukri.org/projects?ref=EP%2FT02450X%2F1},
  note         = {Engineering and Physical Sciences Research Council (EPSRC) Project}
}

\end{document}